\documentclass[letterpaper, 10 pt, conference]{class/ieeeconf}  

\IEEEoverridecommandlockouts                              

\usepackage{siunitx}  
\usepackage{graphicx}

\usepackage{amsmath}
\usepackage{amssymb}
\usepackage{latexsym}
\usepackage{url}
\usepackage{cite}
\usepackage{relsize}
\usepackage{multirow}
\usepackage{afterpage}
\usepackage{ifthen}
\usepackage{graphicx}
\usepackage{algpseudocode,algorithm,algorithmicx}
\usepackage{subfigure}
\usepackage{flushend}
\usepackage{epstopdf}
\usepackage{stackengine}
\usepackage{textcomp} 
\usepackage{gensymb}
\usepackage{booktabs}
\usepackage{makecell}
\usepackage{hhline}
\usepackage{courier}
\usepackage{lipsum}
\usepackage{xspace}
\usepackage{xcolor} 
\usepackage{soul}
\usepackage{amssymb}
\usepackage{pifont}
\usepackage[font=footnotesize,labelfont=bf]{caption}
\usepackage[most]{tcolorbox} 

\makeatletter
\let\NAT@parse\undefined
\makeatother
\usepackage[bookmarks=false, linkcolor=blue, urlcolor=blue, citecolor=blue]{hyperref} 
\hypersetup{
    colorlinks=true,
    linkcolor=red,
    filecolor=magenta,      
    urlcolor=blue,
    pdfstartview={FitH},
    citecolor =blue
    }

\newcommand{\model}{\texttt{SlotVLA}\xspace}

\definecolor{brickred}{rgb}{0.8, 0.25, 0.33}
\definecolor{darkolivegreen}{rgb}{0.33, 0.42, 0.18}

\newcommand{\cmark}{\textcolor{darkolivegreen}{\ding{51}}}
\newcommand{\xmark}{\textcolor{brickred}{\ding{55}}}

\newcommand{\myheading}[1]{\vspace{1ex}\noindent \textbf{#1}}

\usepackage{xspace}
\usepackage{amssymb}
\usepackage{pifont}

 \graphicspath{{./figures/}}
 
\title{\LARGE \bf
\model: Towards Modeling of Object-Relation Representations \\in Robotic Manipulation
}

\author{Taisei Hanyu$^{1,*}$, Nhat Chung$^{2,*,\dagger}$, Huy Le$^{2,*}$, Toan Nguyen$^{2}$, \\
Yuki Ikebe$^{1}$, Anthony Gunderman$^{1}$, Duy Nguyen Ho Minh$^{3,7,8}$, Khoa Vo$^{1}$, \\ Tung Kieu$^{4}$, Kashu Yamazaki$^{5}$, Chase Rainwater$^{1}$, Anh Nguyen$^{6}$, Ngan Le$^{1,\dagger}$
\thanks{$^{*}$These authors contributed equally.} \thanks{$^{\dagger}$Correspondence: {\texttt{nhatcm3@fpt.com}; \texttt{thile@uark.edu}}}%
\thanks{$^{1}$University of Arkansas, USA \quad $^{2}$FPT Software AI Center, Vietnam}%
\thanks{$^{3}$University of Stuttgart, Germany \ \quad $^{4}$Aalborg University, Denmark}%
\thanks{$^{5}$Carnegie Mellon University, USA \quad $^{6}$University of Liverpool, UK}%
\thanks{$^{7}$German Research Center for Artificial Intelligence, Germany} 
\thanks{$^{8}$Max Planck Research School for Intelligent Systems, Germany}%
}

\begin{document}


\newtheorem{problem}{Problem}
\newtheorem{lemma}{Lemma}
\newtheorem{theorem}[lemma]{Theorem}
\newtheorem{claim}{Claim}
\newtheorem{corollary}[lemma]{Corollary}
\newtheorem{definition}[lemma]{Definition}
\newtheorem{proposition}[lemma]{Proposition}
\newtheorem{remark}[lemma]{Remark}
\newenvironment{LabeledProof}[1]{\noindent{\it Proof of #1: }}{\qed}

\def\beq#1\eeq{\begin{equation}#1\end{equation}}
\def\bea#1\eea{\begin{align}#1\end{align}}
\def\beg#1\eeg{\begin{gather}#1\end{gather}}
\def\beqs#1\eeqs{\begin{equation*}#1\end{equation*}}
\def\beas#1\eeas{\begin{align*}#1\end{align*}}
\def\begs#1\eegs{\begin{gather*}#1\end{gather*}}

\newcommand{\poly}{\mathrm{poly}}
\newcommand{\eps}{\epsilon}
\newcommand{\e}{\epsilon}
\newcommand{\polylog}{\mathrm{polylog}}
\newcommand{\rob}[1]{\left( #1 \right)} 
\newcommand{\sqb}[1]{\left[ #1 \right]} 
\newcommand{\cub}[1]{\left\{ #1 \right\} } 
\newcommand{\rb}[1]{\left( #1 \right)} 
\newcommand{\abs}[1]{\left| #1 \right|} 
\newcommand{\zo}{\{0, 1\}}
\newcommand{\zonzo}{\zo^n \to \zo}
\newcommand{\zokzo}{\zo^k \to \zo}
\newcommand{\zot}{\{0,1,2\}}
\newcommand{\en}[1]{\marginpar{\textbf{#1}}}
\newcommand{\efn}[1]{\footnote{\textbf{#1}}}
\newcommand{\vecbm}[1]{\boldmath{#1}} 
\newcommand{\uvec}[1]{\hat{\vec{#1}}}
\newcommand{\thv}{\vecbm{\theta}}
\newcommand{\junk}[1]{}
\newcommand{\var}{\mathop{\mathrm{var}}}
\newcommand{\rank}{\mathop{\mathrm{rank}}}
\newcommand{\diag}{\mathop{\mathrm{diag}}}
\newcommand{\tr}{\mathop{\mathrm{tr}}}
\newcommand{\acos}{\mathop{\mathrm{acos}}}
\newcommand{\atantwo}{\mathop{\mathrm{atan2}}}
\newcommand{\SVD}{\mathop{\mathrm{SVD}}}
\newcommand{\quadf}{\mathop{\mathrm{q}}}
\newcommand{\linterp}{\mathop{\mathrm{l}}}
\newcommand{\sgn}{\mathop{\mathrm{sign}}}
\newcommand{\sym}{\mathop{\mathrm{sym}}}
\newcommand{\avg}{\mathop{\mathrm{avg}}}
\newcommand{\mean}{\mathop{\mathrm{mean}}}
\newcommand{\erf}{\mathop{\mathrm{erf}}}
\newcommand{\grad}{\nabla}
\newcommand{\R}{\mathbb{R}}
\newcommand{\defeq}{\triangleq}
\newcommand{\dims}[2]{[#1\!\times\!#2]}
\newcommand{\sdims}[2]{\mathsmaller{#1\!\times\!#2}}
\newcommand{\udims}[3]{#1}
\newcommand{\udimst}[4]{#1}
\newcommand{\com}[1]{\rhd\text{\emph{#1}}}
\newcommand{\ind}{\hspace{1em}}
\newcommand{\argmin}[1]{\underset{#1}{\operatorname{argmin}}}
\newcommand{\floor}[1]{\left\lfloor{#1}\right\rfloor}
\newcommand{\step}[1]{\vspace{0.5em}\noindent{#1}}
\newcommand{\quat}[1]{\ensuremath{\mathring{\mathbf{#1}}}}
\newcommand{\norm}[1]{\left\lVert#1\right\rVert}
\newcommand{\ignore}[1]{}
\newcommand{\specialcell}[2][c]{\begin{tabular}[#1]{@{}c@{}}#2\end{tabular}}
\newcommand*\Let[2]{\State #1 $\gets$ #2}
\newcommand{\algorithmicbreak}{\textbf{break}}
\newcommand{\Break}{\State \algorithmicbreak}
\newcommand{\ra}[1]{\renewcommand{\arraystretch}{#1}}

\renewcommand{\vec}[1]{\mathbf{#1}} 

\newcommand{\blue}[1]{\textcolor{blue}{#1}}
\newcommand{\red}[1]{\textcolor{red}{#1}}

\maketitle
\thispagestyle{empty}
\pagestyle{empty}

\begin{abstract}

Inspired by how humans reason over discrete objects and their relationships, we explore whether compact object-centric and object-relation representations can form a foundation for multitask robotic manipulation.
Most existing robotic multitask models rely on dense embeddings that entangle both object and background cues, raising concerns about both efficiency and interpretability. 
In contrast, we study object–relation-centric representations as a pathway to more structured, efficient, and explainable visuomotor control.
Our contributions are two-fold. First, we introduce \textbf{LIBERO+}, a fine-grained benchmark dataset designed to enable and evaluate object-relation reasoning in robotic manipulation. Unlike prior datasets, LIBERO+ provides object-centric annotations that enrich demonstrations with box- and mask-level labels as well as instance-level temporal tracking, supporting compact and interpretable visuomotor representations.
Second, we propose \model, a slot-attention–based framework that captures both objects and their relations for action decoding. It uses a slot-based visual tokenizer to maintain consistent temporal object representations, a relation-centric decoder to produce task-relevant embeddings, and an LLM-driven module that translates these embeddings into executable actions.
Experiments on LIBERO+ demonstrate that object-centric slot and object-relation slot representations drastically reduce the number of required visual tokens, while providing competitive generalization. Together, LIBERO+ and \model provide a compact, interpretable, and effective foundation for advancing object–relation-centric robotic manipulation.
\end{abstract}

\section{Introduction}
\label{introduction}
\begin{figure}[!t]
    \centering
    \vspace{-0.5em}
    \includegraphics[width=0.9\linewidth]{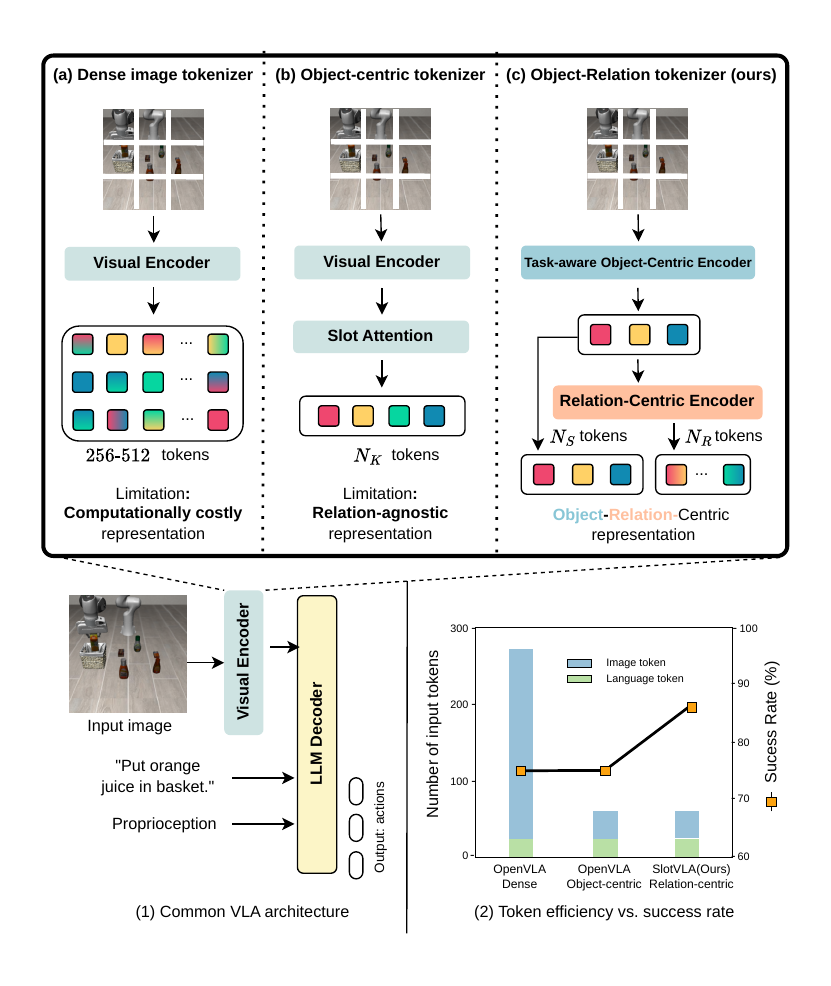}
    \caption{\textbf{Comparison of visuomotor tokenization strategies.} (a) Dense tokenizers generate hundreds of tokens across the scene, leading to computationally costly representations. (b) Object-centric tokenizer yields $N_K$ tokens, each representing an object. (c) Our object–relation-centric tokenizer yields $N_S$ object tokens and $N_R$ relation tokens, producing structured and efficient representations. Plot (2) shows that our method achieves higher success rates with fewer tokens compared to baselines on LIBERO-Goal.
    } 
    \vspace{-0.5em} 
    \label{fig:teaser}
\end{figure}

Recent advances in vision-language-action (VLA) modeling have significantly improved visuomotor control in robotics~\cite{DBLP:OpenVLA2024, black2024pi0, zawalski2024robotic, wang2024hpt},
integrating language conditions with visual cues to enable precise, multitask action prediction across numerous applications~\cite{ma2024survey}.
While many architectures such as OpenVLA~\cite{DBLP:OpenVLA2024}, $\pi_0$~\cite{black2024pi0}, ECoT~\cite{zawalski2024robotic}, HPTs~\cite{wang2024hpt} have contributed to VLA pipelines, even employing a Large Language Model (LLM) for action decoding~\cite{DBLP:OpenVLA2024, zawalski2024robotic}, the vision encoder remains a critical bottleneck, as it serves as the perceptual foundation for action reasoning. In particular, pretrained encoders such as DINOv2~\cite{oquab2023dinov2}, SigLIP~\cite{siglip2023} are widely adopted in VLAs to produce a large number of visual tokens (e.g., 256 to 512), whose computational costs can become increasingly prohibitive as the number of visual tokens grows (as in Fig.~\ref{fig:teaser}a).
While these embeddings are rich in information, they can entangle various information and even redundant background features, limiting interpretability and potentially obscuring task-relevant cues from the action decoder~\cite{tian2024tokenize}. 

Object-centric learning in computer vision has shown remarkable success in producing disentangled, interpretable representations that support generalization across tasks~\cite{jiang2024slot, actionslot2024, zhu2023viola, DBLP:conf/icra/ShiQMJ24,uno}. While promising, directly transferring object-centric methods to robotics is suboptimal due to the interactive nature of embodied manipulation tasks (shown in Fig. \ref{fig:teaser}b of limited relational modeling among object slots).
In robotic environments, scenes are often cluttered, and many objects are irrelevant to the current task. Simply increasing the number of slots does not help, as irrelevant objects or background elements may dominate the representations, preventing slots from consistently capturing meaningful entities \cite{DBLP:conf/nips/LocatelloWUMHUD20, heravi2022visuomotor} and confusing the action decoder. 
Therefore, instead of modeling all objects in the scene, the model must identify and focus on the task-relevant objects that the manipulator interacts with. 
Moreover, purely object-centric encodings fail to capture essential relational cues, most notably the gripper–object interactions. As a result, action decoders are forced to infer control parameters from incomplete relational signals. 
Unlike passive perception, robotic manipulation requires explicit modeling of both objects and their relations to the manipulator and surrounding scene~\cite{cai2024spatialbot, zawalski2024robotic}.

To facilitate the study and future research of object-relation representations for multitask robotic manipulation, we introduce \textbf{LIBERO+}, a fine-grained benchmark that extends LIBERO \cite{libero2023} with explicit object-focused annotations. Originally, while LIBERO provides a diverse set of manipulation tasks, like many existing datasets, it does not include detailed grounding at the object level. LIBERO+ fills this gap by adding box- and mask-level labels along with temporal tracking across RGB-D modalities, enriching demonstrations with structured information that supports object–relation reasoning. These annotations enable systematic design and evaluation of object–relation-centric visuomotor models and open opportunities for interpretable learning.

To address the modeling gap, we propose an object-relation-centric paradigm for robotic VLA models. Concretely, we introduce \textbf{\model}, a framework that employs slot attention with a task-aware filter to extract only relevant object representations and a relation encoder to capture their interactions, enabling relational reasoning under strict token budgets. As shown in Fig.~\ref{fig:teaser}c, the \textbf{\model} framework relies on an Object-centric Encoder that disentangles object features into a compact set of slots and filters them for task relevance, yielding concise and interpretable representations. The Relation Encoder then models their interactions, including those involving the robot gripper and the background context. Finally, both object- and relation-centric tokens are decoded into precise control actions.


\noindent In summary, our contributions are two-fold:
\begin{itemize}
\item \textbf{LIBERO+} dataset: a fine-grained benchmark emphasizing object–relation reasoning with RGB-D input and instance-level annotations for robotic manipulation.
\item \textbf{\model}: an object-relation-centric VLA framework that combines object-centric slots with relation-centric tokens. The object slots capture disentangled entities from the environment and are filtered for task relevance, while the relation tokens (e.g., gripper-object interactions) encode task-aware interactions. Together, they yield a compact and interpretable representation for action decoding.
\end{itemize}

\section{Related Works}
\label{related}
\myheading{VLA Learning in Robotic Manipulation.}
VLA learning has emerged as a powerful paradigm for instruction-following agents in various embodied tasks, including 3D scene reconstruction, navigation, cross-embodiment transfer, and, most notably, robotic manipulation ~\cite{DBLP:OpenVLA2024, black2024pi0, zawalski2024robotic, wang2024hpt}.
As visual perception is a critical bottleneck, many recent architectures such as OpenVLA~\cite{DBLP:OpenVLA2024}, $\pi_0$~\cite{black2024pi0}, ECoT~\cite{zawalski2024robotic}, HPTs~\cite{wang2024hpt} have leveraged strong pretrained vision encoders, like DINOv2~\cite{oquab2023dinov2}, SigLIP~\cite{siglip2023}, that are effective at capturing diverse features. However, they can generate dense representations that intermingle object positions, affordances, and backgrounds, making it challenging for action decoders to isolate task-relevant signals~\cite{DBLP:conf/icra/LinZSIL20}.
Diverging from the recent VLA literature, our research considers a novel low-token perspective towards performing manipulation tasks. {Inspired by the remarkable success of object-centric learning in computer vision~\cite{jiang2024slot, actionslot2024}, we explore
whether a few rich semantic slots (i.e., object slots and object-relation slots) can offer a more efficient and interpretable foundation for robotic manipulation while isolating task-relevant objects out of the redundant background.}

\myheading{Vision Token Reduction.}
Vision-language models (VLM) process numerous tokens, especially in multi-view \cite{DBLP:OpenVLA2024} and video reasoning~\cite{tian2024tokenize}. To reduce memory and computational costs, several token compression methods have been introduced. Approaches such as Token Merging~\cite{tran2025accelerating}, PruMerge~\cite{shang2024llava}, and TokenPacker~\cite{li2024tokenpacker} reduce redundancy by aggregating tokens, while models such as Qwen-VL~\cite{bai2023qwen} and MQT-LLaVA~\cite{hu2024matryoshka} employ Q-Former~\cite{li2023blip} or resampler modules~\cite{wang2024hpt} to construct fixed-length visual representations. However, these methods emphasize general-purpose compression rather than extracting task-relevant and interpretable structures, limiting their applicability to robotics.
Meanwhile, slot-based learning has recently been leveraged to focus on modular, object-centric structures in visual reasoning~\cite{actionslot2024, jiang2024slot, DBLP:journals/corr/abs-2407-06871, vo2025henasy}, 
using just a few semantic slots. 
\textit{Unlike existing works, our \model extends the slot attention mechanism to robot manipulation tasks, particularly by improving upon object-centric slots to form object-scene interaction semantics, preserving their relational features for multitask manipulation control.}


\myheading{Object-Centric Representation in Robot Manipulation}
This strategy is often realized by explicitly encoding objects as bounding boxes \cite{devin2018deep}, keypoints, or poses \cite{tyree20226}, using grounding techniques to feed these tokens into the model \cite{li2025controlvla}. More recently, object-centric information has also been expressed in text prompts with explicit coordinates \cite{wen2024object}, guiding the model’s attention toward task-relevant entities through multimodal grounding. While this provides strong supervision and explicit grounding, it ties performance to the quality of external annotations and limits flexibility in capturing unannotated or relational aspects of the scene \cite{agostini2020efficient,goodwin2022semantically}. In contrast, slot attention discovers object slots directly from raw inputs, enabling flexible and generalizable object representations. \textit{Building on this, our SlotVLA formulation unifies object-centric slots with relation-centric tokens, where slots capture disentangled entities filtered for task relevance, and relation tokens encode task-aware interactions (e.g., gripper–object). This yields a compact, interpretable representation that enhances robot action decoding.}

\section{LIBERO+: Data Curation}
\label{sec:libero-dataset-curation}
\begin{figure}[!t]
    \centering
\includegraphics[width=1.0\linewidth]{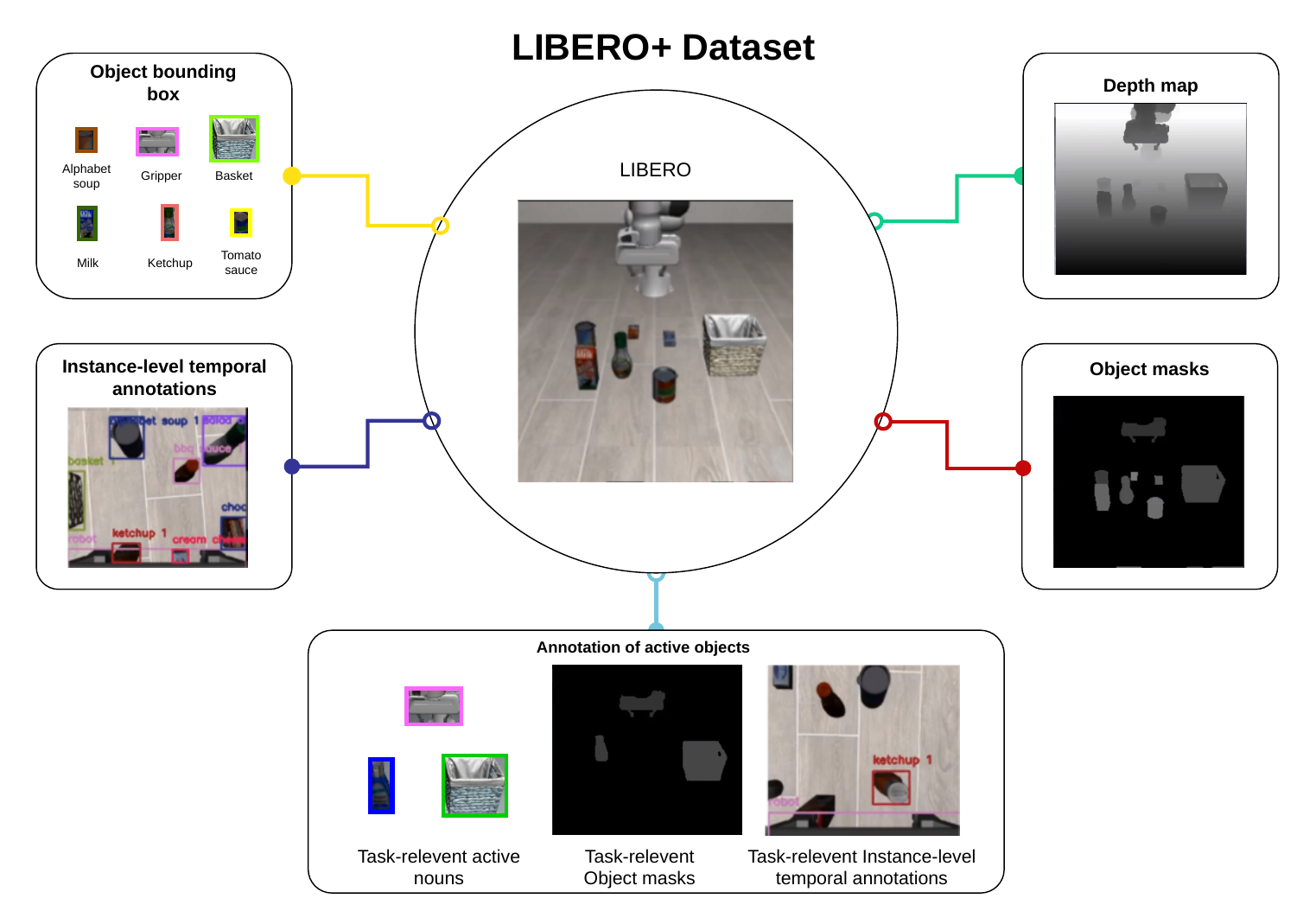}
    \vspace{0.05em} 
    \caption{Overview of the LIBERO+ dataset.} 
    \label{fig:libero_plus}
\end{figure}

We build LIBERO+ as an extension of the LIBERO benchmark suite~\cite{libero2023}. While LIBERO is well-suited for evaluating high-level visuomotor learning, it lacks explicit object-level supervision. To address this gap, LIBERO+ introduces finer-grained, object-centric annotations designed to support compact, interpretable, and low-token VLA representations. As LIBERO is a task suite built on top of robosuite~\cite{robosuite2020}, a non-trivial challenge lies not only in grounding natural language descriptions to object entities in the simulator, but also in representing each object holistically as a single “object” is often composed of multiple underlying assets (e.g., meshes, textures, physical handles), which are fragmented into disjointed parts, reducing both interpretability and utility for fine-grained visuomotor reasoning. To address this, we manually align low-level asset names with specific objects and label their natural linguistic references, ensuring consistency across entire trajectories. In addition, we unify semantic object masks and disambiguate multiple instances of the same object by leveraging preassigned asset names, producing coherent and interpretable masks that also form bounding boxes. Then, in order to provide task-relevant, active objects, we include task-specific naturalistic nouns for each demonstration to filter for relevant object labels.  Specifically, as seen in Fig.~\ref{fig:libero_plus}, LIBERO+ augments the original demonstrations with both \textbf{box-level} and \textbf{mask-level} object annotations, as well as \textbf{instance-level temporal ID tracking} across both RGB and Depth modalities, as follows:
\vspace{-0.1in}
\begin{tcolorbox}[colback=gray!10,colframe=black]
\noindent
$\bullet$ \texttt{Bounding boxes} provide 2D spatial anchors that localize objects, serving as entry points for token extraction.

\noindent
$\bullet$ \texttt{Object masks} offer pixel-level segmentations aligned with bounding boxes, preserving object boundaries and preventing feature entanglement with background pixels.

\noindent
$\bullet$ \texttt{Instance-level temporal IDs} maintain consistent object identities across frames (e.g., basket1, plate1, plate2), enabling objects to be temporally tracked throughout a sequence and supporting long-horizon reasoning. 

\noindent
$\bullet$ \texttt{Depth map}, restricted to masked regions, provides per-pixel depth signals that encode occlusions and relative distances, which are critical for gripper–object reasoning.

\noindent 
$\bullet$ \texttt{Task-relevant objects} provide the set of objects explicitly mentioned in the task description. For example, given the task "robot put the bowl on top of the cabinet", the identified task-relevant objects are robot, bowl, and cabinet.
\end{tcolorbox}

LIBERO+ includes four subsets, derived from the original LIBERO suite: LIBERO-Goal, LIBERO-Object, LIBERO-Spatial, and LIBERO-Long. Each subset is curated to emphasize different aspects of manipulation, while consistently providing structured object–relation annotations to support fine-grained reasoning. For temporally consistent action supervision, we retain LIBERO's native action labels but introduce a filtering step to remove redundant \textit{no-op} actions. This refinement reduces idle-frame redundancy and sharpens the alignment between annotated objects and action-relevant dynamics. As a result, LIBERO+ yields compact yet semantically rich representations, where object-centric slots are directly grounded in action-relevant cues. This design ensures that LIBERO+ serves as a challenging and comprehensive testbed for advancing efficient, interpretable, and object-relation-centric VLA reasoning.

The statistical summary of LIBERO+, constructed from four subsets: LIBERO-Object (L-Object), LIBERO-Goal (L-Goal), LIBERO-Spatial (L-Spatial), and LIBERO-Long (L-Long), is presented in Table~\ref{tab:libero_stats}.


\vspace{0.5em} 
\begin{table}[h]
\centering
\caption{Statistics of LIBERO+. TR corresponds to Task-relevant objects.}
\vspace{1em}
\resizebox{0.85\linewidth}{!}{\begin{tabular}{lcccc}
\toprule
\multirow{2}{*}{\textbf{Statistics}}&  \multicolumn{4}{c}{ \textbf{LIBERO+} }\\ \cmidrule{2-5}
 & {L-Object} & {L-Goal} & {L-Spatial} & {L-Long} \\
\toprule
\# Tasks              & 10       & 10       & 10       & 10 \\
\# Object Layouts            & 1        & 1        & 10       & 9   \\
\# Objects            & 12       & 7        & 11       & 29  \\
\# TR Objects  & 2--3        & 2--3        & 3--4        & 3--4   \\
\# Total Frames           & 72,063       & 54,779       & 47,253   &     84,896  \\
\# Total BBoxes          & 570,328  & 374,692  & 510,985  & 487,333  \\
\# TR BBoxes           & 285,912  & 130,814  & 221,468  & 257,105 \\
\bottomrule
\end{tabular}}
\label{tab:libero_stats}
\end{table}

\vspace{-0.1in}
\section{Methodology}
\vspace{-0.1in}
\subsection{Overview}
The goal of \model is to achieve token-efficient visuomotor reasoning by transforming dense visual embeddings into compact object–relation representations. Conventional encoders produce hundreds of tokens (typically 256–512), which are computationally expensive for downstream reasoning. Object-centric representations address this overhead by focusing on discrete objects, but they overlook relational cues such as gripper–object interactions.  

\begin{figure*}[t!]
\vspace{0.1cm}
\begin{center}
\centerline{\includegraphics[width=0.95\linewidth]{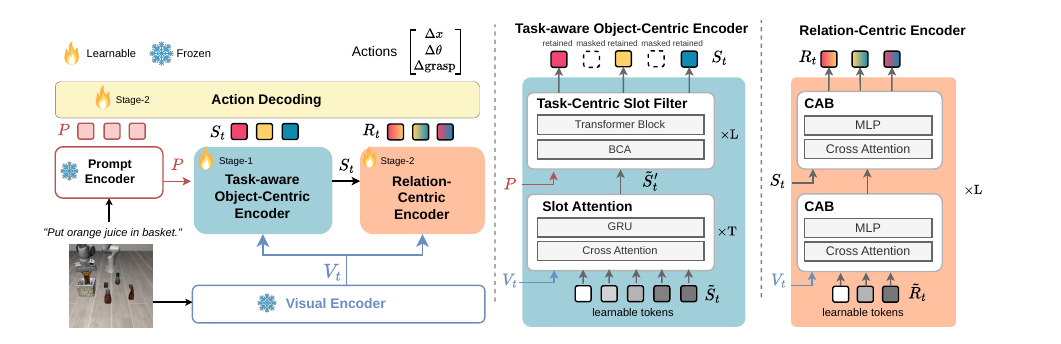}}
\vskip 0.05in
\caption{Overall framework of our proposed model. Stage-1 trains the Task-aware Object-Centric Encoder with slot attention and task-aware filtering. Stage-2 freezes Stage-1 parameters and introduces the Relation-Centric Encoder, enabling relational reasoning for final action decoding. }
\label{fig:slotvla}
\end{center}
\vskip -0.2in
\end{figure*}

In contrast, our approach compresses the dense input into object-centric representations and applies a task-aware filtering mechanism that discards irrelevant objects in the environment. This yields a clearer representation with only a few slots (around 4), corresponding to the 2–4 task-relevant objects observed across LIBERO tasks (Table~\ref{tab:libero_stats}). Additionally, the object slots are augmented with a small number of relation tokens to explicitly capture interactions among objects and with the manipulator, resulting in a representation that is both compact and relationally expressive. The overall architecture of this framework is illustrated in Fig.~\ref{fig:slotvla}.


\subsection{Problem Formulation}
\label{sec:prob-form}

Formally, let $\mathbf{V}_t = \{ \mathbf{v}_t^1, \dots, \mathbf{v}_t^N \}$ denote a $N$ dense set of visual tokens/patches from the feature map encoded by a visual encoder from the image $\mathbf{I}_t$ at time $t$ and $\mathbf{P} = \{\mathbf{p}^1, \dots, \mathbf{p}^M\}$ represents the embeddings of language tokens of task description. Our goal is to learn a function $g_\phi(\cdot)$ that gives semantically-rich compact representation as:
\begin{equation}
\{ \mathbf{S}_t , \mathbf{R}_t  \} = g_\phi(\mathbf{V}_t, \mathbf{P})
\end{equation}
\noindent where $\mathbf{S}_t \in \mathbb{R}^{N_S \times d}$ denotes the object-centric representations with $N_S$ tokens, and $\mathbf{R}_t \in \mathbb{R}^{N_R \times d}$ denotes the relation-centric representations that capture interactions among slots and with dense features using $N_R$ tokens. Importantly, $N_S + N_R \ll N$, enabling downstream reasoning tasks to be carried out in a token-efficient manner.
Given such visual token representations along with task embedding $\mathbf{P}$ and proprioception embedding $\mathbf{o}_t$, our final goal is to train an action decoding function ${f}_\theta(\cdot)$ that predicts action logits $A_t$ over possible actions that a robot can take to solve the task:
\begin{equation}
A_t = f_\theta(\mathbf{S}_t , \mathbf{R}_t, \mathbf{P}, \mathbf{o}_t)
\end{equation}

\subsection{Task-Aware Object-Centric Encoder}
\label{subsec:slot-tokenizer}

As task scaling requires open-vocabulary object extraction at test time, we aim to reduce redundancy in dense tokens by emphasizing task-relevant representations. To this end, we design a slot-based encoder that compresses dense visual tokens into compact object-centric tokens, guided by language-based task filtering. The encoder comprises two modules: \textit{(1) slot attention with temporal consistency}, which extracts object-centric slots from dense features and maintains their updates over time; and \textit{(2) task-aware slot filtering}, which selects the slots most relevant to the manipulation task.

\subsubsection{Slot Attention with Temporal Consistency}
On top of the visual encoder, we employ slot attention~\cite{DBLP:conf/nips/LocatelloWUMHUD20} to map dense visual patches $\mathbf{V}_t \in \mathbb{R}^{N\times d}$ into a set of learnable slots 
$\Tilde{\mathbf{S}}_t \in \mathbb{R}^{N_{\Tilde{S}}\times d}$. This iterative attention process, implemented with a GRU~\cite{DBLP:conf/emnlp/ChoMGBBSB14}, captures modular semantics~\cite{xu2022groupvit,DBLP:conf/iclr/JiaLH23} and produces object-centric tokens:
\begin{align}
    \Tilde{\mathbf{a}}_{i,j} &= \frac{e^{\mathbf{a}_{i,j}}}{\sum_{l} e^{\mathbf{a}_{i,l}}},
    \quad \text{where} \quad
    \mathbf{a} = \tfrac{1}{\sqrt{d}}k(\mathbf{V}_t)q(\Tilde{\mathbf{S}})^{\top} \notag \\
    \mathbf{w}_{i,j} &= \frac{\Tilde{\mathbf{a}}_{i,j}}{\sum_{l}\Tilde{\mathbf{a}}_{l,j}} \notag \\
    \Tilde{\mathbf{S}}'_t &= \text{GRU}(\texttt{inputs}=\mathbf{w^\top}v(\mathbf{V}_t),
    \texttt{states}=\Tilde{\mathbf{S}}_t). \label{eq:gru_update}
\end{align}
\noindent where linear transformation heads $q(\cdot)$, $k(\cdot)$, $v(\cdot)$ are used to map learnable slots $\Tilde{\mathbf{S}}_t$ and frame-wise feature maps $\mathbf{V}_t$. 

To ensure \textit{temporal consistency} in object identity, slots are initialized through a \emph{slot carryover mechanism},
\begin{equation}
\label{eq:slot_init}
   \Tilde{\mathbf{S}}_t^{(0)} = 
    \begin{cases}
        \text{RandomInit}(), & t = 0 \\
        \Tilde{\mathbf{S}}_{t-1}^{(T)}, & t > 0,
    \end{cases}
\end{equation}
where $T$ is the number of refinement steps per frame. Thus, slots start randomly at $t=0$ and are propagated from the previous timestep otherwise. 

\begin{figure*}[t!]
\vspace{0.1cm}
\begin{center}
\centerline{\includegraphics[width=1.0\linewidth]{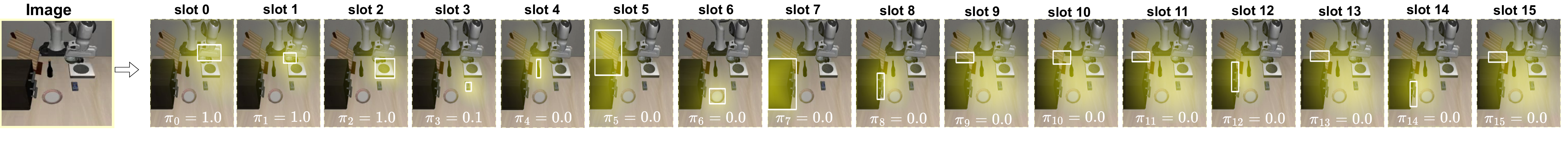}}
\vskip 0.05in
\caption{Slot decomposition result. Task query: ``\textit{Put the bowl on the stove}''. Task-relevant slots correctly bind to objects, while irrelevant slots scatter.}
\label{fig:slot_decomposition}
\end{center}
\vskip -0.3in
\end{figure*}

\subsubsection{Task-Aware Slot Filter}
\label{subsec:task-centric}
While leveraging slots is practical for robotic manipulation, not all slots are relevant to a given task. We therefore introduce a \emph{task-aware slot filter} to select $N_S$ slots from $N_{\Tilde{S}}$ using bidirectional cross-attention ($\operatorname{BCA}$) \cite{li2022grounded}, followed by a transformer layer ($\operatorname{Trans}$), by estimating the relevance scores of the tokens with respect to the task description. Specifically, the task-aware slot filter computes the relevant scores $\pi_t$ at time $t$ between the object-centric slots $\Tilde{\mathbf{S}}'_t$ from the GRU and the embeddings of language tokens $\mathbf{P}$:
\begin{equation}
\label{eq:relevent_score}
{\pi}_t = \operatorname{Trans}(\operatorname{BCA}(\Tilde{\mathbf{S}}', \mathbf{P})).
\end{equation}

Then, we retain $N_S$ most relevant slots during training and inference with a ${\operatorname{Top}_{k}}(\cdot)$ function and filter out the rest. Given the the object-centric slots $\Tilde{\mathbf{S}}'_t$, we extract a refined subset $\mathbf{S}_t=\{\tilde{\mathbf{s}}_t^1, \ldots, \tilde{\mathbf{s}}_t^{N_S}\}$ to serve as task-aware object-centric tokens, where $N_S \leq N_{\Tilde{S}}$, based on their scores:
\vspace{-0.2em}
\begin{equation}
\mathbf{S}_t = \{\mathbf{\Tilde{s}}^i_t \mid i \in \operatorname{Top}_k(\pi_t) \}.
\end{equation}

\subsection{Relation-Centric Encoder}
\label{subsec:relation_centric_enc}
In robotic tasks, understanding of interactions between objects is crucial for planning grasps and manipulations. Although object-centric tokens localize items of interest, they do not inherently capture relationships among objects (e.g., the gripper and objects). 
Specifically, we model relations using learnable queries, $\Tilde{\mathbf{R}}_t$. These queries integrate information from dense visual patches ($\mathbf{V}_t$) and object-centric tokens ($\mathbf{S}_t$) using a Cross-Attention Block (CAB), which consists of multi-head cross-attention and feed-forward layers. The CAB applies visual conditioning to process the patches and slot-based conditioning to process the object tokens.
\vspace{-0.5em}
\begin{equation}
\mathbf{R}_t = \operatorname{CAB}(\operatorname{CAB}(\Tilde{\mathbf{R}}_t, \mathbf{V}_t), \mathbf{S}_t).
\end{equation}






\subsection{Action Decoding}
\label{subsec:action_dec}
Inspired by~\cite{tian2024tokenize}, we employ LoRA~\cite{DBLP:conf/iclr/LoRA2022} to integrate the object and relation tokens into a VLA framework that leverages an LLM action decoder, which we denote as $\operatorname{AD}_{\theta_{\text{LoRA}}}(\cdot)$. The task-aware object-centric tokens $\mathbf{S}_t$ and the relation tokens $\mathbf{R}_t$
are concatenated together with the language embeddings $\mathbf{P}$ from the task description and the robot proprioception $\mathbf{o}_t$ to form a multimodal input sequence. Then, we define the action prediction as a classification problem, where each control dimension 
is discretized via binning \cite{DBLP:OpenVLA2024} or tokenization \cite{pertsch2025fast}. 

\vspace{-0.5em}
\begin{equation}
a_t = \underset{k}{\operatorname{argmax}} \operatorname{AD}_{\theta_{\text{LoRA}}}([\mathbf{S}_t \ ; \mathbf{R}_t \ ; \mathbf{P} \ ; \mathbf{o}_t] )
\vspace{-0.5em}
\end{equation}

\noindent where $[;]$ denotes the concatenation operation, and the action $a_t$ is obtained by greedy decoding over action logits $A_t$. Recall that $N_S + N_R \ll N$; thus, this approach is much token efficient than using  $[\mathbf{V}_t; \mathbf{P}, \mathbf{o}_t]$ as input.





\subsection{Objective Functions}
We adopt a two-stage training strategy: first, we supervise the Task-aware Object-Centric Encoder (Section \ref{subsec:slot-tokenizer}), and then we fine-tune the Relation-Centric Encoder (Section \ref{subsec:relation_centric_enc}) jointly with the Action Decoding (Section \ref{subsec:action_dec}).

\subsubsection{Stage -- 1 Training} We first train the Task-aware Object-Centric
Encoder with two objectives that include: (a) slot-attention supervision and (b) task-aware slot filter supervision (Fig. \ref{fig:slotvla}: Stage-1).
\begin{equation}
\mathcal{L}_{\text{slot-enc}} \;=\;
\lambda_{\text{slot-attn}}\,\mathcal{L}_{\text{slot-attn}}
\;+\;
\lambda_{\text{track}}\,\mathcal{L}_{\text{track}}
\;+\;
\lambda_{\text{int}}\,\mathcal{L}_{\text{int}} .
\end{equation}
where, given $N_{\tilde S}$ predicted slots and $N_G$ ground-truth objects per frame with the requirement of $N_G < N_{\tilde S}$, we align predicted slots with ground-truth objects at each timestep via Hungarian  matching~\cite{kuhn1955hungarian} using box-based costs. 

\noindent\emph{(a) Slot Attention Supervision:}
We supervise the slot attention with the bounding boxes, objectness, and masks:
\begin{equation}
\mathcal{L}_{\text{slot-attn}} \;=\;
\lambda_{\text{box}}\,\mathcal{L}_{\text{box}}
\;+\;
\lambda_{\text{obj}}\,\mathcal{L}_{\text{obj}}
\;+\;
\lambda_{\text{seg}}\,\mathcal{L}_{\text{seg}}.
\end{equation}
Here, $\mathcal{L}_{\text{box}}$ is a standard DETR-style box loss \cite{carion2020end}, $\mathcal{L}_{\text{obj}}$ is a binary cross-entropy (BCE) loss that assigns $1$ to matched slots and $0$ to unmatched ones, thereby encouraging slots to activate only when corresponding to real objects. $\mathcal{L}_{\text{seg}}$ is a pixel-wise BCE that enforces finer alignment between predicted and ground-truth instance masks. Note that the supervising signals are coming from our introduced LIBERO+ dataset. Then, we enforce temporal consistency with a tracking loss, aligning slots with the same object across frames using bounding box and mask annotations:
\begin{equation}
\label{eq:contrastive_loss_short}
\mathcal{L}_{\text{track}}
=
-\sum_{(i,t)}
\log
\frac{
\sum\limits_{(i',t')\in \mathcal{P}(i,t)}
\exp\!\left({\mathrm{sim}(\mathbf{s}_t^{i}, \mathbf{s}_{t'}^{i'})}/{\tau}\right)
}{
\sum\limits_{(j,t'')\in \mathcal{P}(i,t)\cup \mathcal{N}(i,t)}
\exp\!\left({\mathrm{sim}(\mathbf{s}_t^{i}, \mathbf{s}_{t''}^{j})}/{\tau}\right)
},
\end{equation}
with $\mathbf{s}_t^i = (\tilde{\mathbf{S}}_t^{(T)})_i$, where $\mathcal{P}(i,t)$ denotes slots of the same object across nearby frames (positives), and $\mathcal{N}(i,t)$ denotes slots from other objects or different videos (negatives).



\noindent\emph{(b) Task-Aware Slot Filter Supervision:}
We directly supervise the task relevence score $\pi_t^i$ for each slot $i$ from Eq. \ref{eq:relevent_score} with a class-imbalanced BCE:
\begin{equation}
\mathcal{L}_{\text{int}} =
\frac{1}{N}\sum_{i,t} w(\hat{u}_{i, t}) \,\text{BCE}(\hat{u}_{i,t}, \pi_t^i),
\end{equation}
where $\hat{u}_{i,t}\!\in\!\{0,1\}$ indicates whether slot $i$ is relevant to the instruction at time $t$. The weight function $w(\cdot)$ up-weights positive labels to address class imbalance 
(e.g., $w(1)=2.0$, $w(0)=1.0$ in our experiments).

\subsubsection{Stage -- 2 Training} 
We used cross-entropy (CE) loss between the predicted action logits $A_t$ before the argmax operation and the ground-truth one-hot action label $\hat{A}_t$ to train the Action Decoder together with Relation-Centric Encoder (Fig.~\ref{fig:slotvla}: Stage-2). 
\vspace{-0.5em}
\begin{equation}
\mathcal{L}_{CE} 
= 
- \sum_{t=1}^{L} 
\hat{A}_t \,\log A_t,
\label{eq:ce_loss}
\end{equation}
where $L$ is the total number of action steps. Note that Stage-1 parameters are frozen at this stage.


\section{EXPERIMENTS}
\subsection{Benchmarks and Evaluation Strategies} 

We conduct experiments on LIBERO+ benchmark (see details in Section~\ref{sec:libero-dataset-curation}). The diversity across its four subsets---in terms of layouts, object counts, and task complexity---enables us to evaluate the models' ability to balance spatial precision, relational reasoning, and temporal consistency.

\begin{table*}[!h]
\centering
\setlength{\tabcolsep}{4pt}
\vspace{1em}
\caption{Benchmark on LIBERO+ consisting of four subsets from LIBERO. Highest results are \textbf{bolded}, second-highest are \underline{underlined}. No. Token indicates the number of tokens used. The tasks, numbered from 1 to 10, are specific to each subset and sorted alphabetically.}
\vspace{1em}
\resizebox{\linewidth}{!}{
\begin{tabular}{l|ccc|ccc|ccc|ccc}
\toprule
       & \multicolumn{3}{c|}{LIBERO-Goal}                   & \multicolumn{3}{c|}{LIBERO-Spatial}                     & \multicolumn{3}{c|}{LIBERO-Object}                  & \multicolumn{3}{c}{LIBERO-Long}       
       \\ 
       & OpenVLA & OC & ORC & OpenVLA & OC & ORC & OpenVLA & OC & ORC & OpenVLA & OC & ORC\\ \toprule
\textbf{Average} & \underline{0.77} & \underline{0.77} & \textbf{0.86} & \textbf{0.72} & 0.48 & \underline{0.60} & 0.70 & \underline{0.90} & \textbf{0.91} & \textbf{0.56} & 0.12 & \underline{0.31} \\  \midrule
\textbf{No. Token} & 256 & 4 (\textcolor{green}{64$\times$}) & 20 (\textcolor{green}{13$\times$})
                   & 256 & 4 (\textcolor{green}{64$\times$}) & 28 (\textcolor{green}{9$\times$})
                   & 256 & 4 (\textcolor{green}{64$\times$}) & 28 (\textcolor{green}{9$\times$})
                   & 256 & 4 (\textcolor{green}{64$\times$}) & 28 (\textcolor{green}{9$\times$}) \\
\textbf{GFLOPs}    & 2{,}112 & 561 (\textcolor{green}{4$\times$}) & 697 (\textcolor{green}{3$\times$})
                   & 2{,}112 & 568 (\textcolor{green}{4$\times$}) & 723 (\textcolor{green}{3$\times$})
                   & 2{,}112 & 568 (\textcolor{green}{4$\times$}) & 723 (\textcolor{green}{3$\times$})
                   & 2{,}112 & 568 (\textcolor{green}{4$\times$}) & 723 (\textcolor{green}{3$\times$}) \\
\midrule
Task 1  & 0.60 & 0.90 & 0.70 & 0.82 & 0.70 & 0.65 & 0.75 & 0.85 & 0.95 & 0.75 & 0.50 & 0.20 \\ 
Task 2  & 0.95 & 0.50 & 0.75 & 0.95 & 0.90 & 0.20 & 0.70 & 0.80 & 0.80 & 0.90 & 0.00 & 0.00 \\ 
Task 3  & 0.70 & 0.90 & 1.00 & 0.82 & 0.00 & 0.40 & 0.85 & 1.00 & 1.00 & 0.55 & 0.10 & 0.40 \\ 
Task 4  & 0.50 & 0.75 & 1.00 & 0.92 & 0.75 & 0.85 & 0.45 & 1.00 & 1.00 & 0.55 & 0.10 & 0.15 \\ 
Task 5  & 0.95 & 1.00 & 0.95 & 0.70 & 0.80 & 0.90 & 0.95 & 1.00 & 0.70 & 0.40 & 0.15 & 0.40 \\ 
Task 6  & 0.90 & 0.95 & 1.00 & 0.85 & 0.70 & 0.75 & 0.60 & 0.95 & 0.95 & 0.75 & 0.05 & 0.65 \\ 
Task 7  & 0.75 & 0.35 & 0.85 & 0.88 & 0.70 & 0.65 & 0.45 & 0.90 & 1.00 & 0.40 & 0.10 & 0.55 \\ 
Task 8  & 0.90 & 1.00 & 0.50 & 0.73 & 0.15 & 0.45 & 0.80 & 0.95 & 0.85 & 0.60 & 0.05 & 0.20 \\ 
Task 9  & 0.90 & 0.60 & 0.90 & 0.82 & 0.00 & 0.80 & 0.50 & 0.85 & 0.95 & 0.35 & 0.00 & 0.10 \\ 
Task 10 & 0.52 & 0.85 & 1.00 & 0.62 & 0.00 & 0.35 & 0.70 & 0.70 & 0.90 & 0.40 & 0.15 & 0.40 \\  \bottomrule        
\end{tabular}}
\label{tab:main_results}
\end{table*}

\begin{table*}[]
\centering
\setlength{\tabcolsep}{9pt}
\caption{Ablation study on Task-Aware Slot Filtering. Object-centric slots (OC) and object–relation-centric slots (ORC) are compared. \cmark indicates that filtering is included, while \xmark indicates that filtering is not included. The tasks, numbered from 1 to 10 are specific to each subset and sorted alphabetically.}
\vspace{1em}
\resizebox{\linewidth}{!}{ \begin{tabular}{l|cccc|cccc|cccc|cccc}
\toprule
       & \multicolumn{4}{c|}{LIBERO-Goal}                   & \multicolumn{4}{c|}{LIBERO-Spatial}                     & \multicolumn{4}{c|}{LIBERO-Object}                  & \multicolumn{4}{c}{LIBERO-Long} \\ \cmidrule{2-17}
       & \multicolumn{2}{c}{OC}   & \multicolumn{2}{c|}{ORC} & \multicolumn{2}{c}{OC}   & \multicolumn{2}{c|}{ORC} & \multicolumn{2}{c}{OC}   & \multicolumn{2}{c|}{ORC} & \multicolumn{2}{c}{OC}   & \multicolumn{2}{c}{ORC} \\ \cmidrule{2-17}
       \textbf{Language}& \xmark & \cmark & \xmark & \cmark & \xmark & \cmark & \xmark & \cmark & \xmark & \cmark & \xmark & \cmark & \xmark & \cmark & \xmark & \cmark \\ \midrule
\textbf{Average} & \underline{0.77} & \underline{0.77} & 0.72 & \textbf{0.86} & \underline{0.53} & 0.48 & \textbf{0.60} & \textbf{0.60} & 0.76 & \underline{0.90} & \textbf{0.91} & \textbf{0.91} & {0.11} & 0.07 & \underline{0.12} & \textbf{0.31} \\ \midrule
{Task 1}  & 0.75 & {0.90} & 0.60 & {0.70} & {0.75} & 0.70 & {0.80} & 0.65 & {0.95} & 0.85 & 0.75 & {0.95} & 0.05 & {0.15} & 0.50 & {0.20} \\ 
{Task 2}  & {1.00} & 0.50 & 0.40 & {0.75} & 0.70 & {0.90} & {0.20} & {0.20} & 0.40 & {0.80} & {0.85} & 0.80 & {0.00} & {0.00} & {0.00} & {0.00} \\ 
{Task 3}  & {0.95} & 0.90 & 0.70 & {1.00} & {0.15} & 0.00 & {0.75} & 0.40 & {1.00} & {1.00} & 0.85 & {1.00} & {0.25} & 0.00 & 0.10 & {0.40} \\ 
{Task 4}  & 0.60 & {0.75} & 0.75 & {1.00} & 0.20 & {0.75} & {0.95} & 0.85 & {1.00} & {1.00} & {1.00} & {1.00} & {0.05} & 0.00 & 0.10 & {0.15} \\ 
{Task 5}  & {1.00} & {1.00} & {1.00} & 0.95 & {0.90} & 0.80 & 0.70 & {0.90} & 0.85 & {1.00} & {0.85} & 0.70 & 0.10 & {0.15} & 0.15 & {0.40} \\ 
Task 6  & 0.90 & {0.95} & 0.85 & {1.00} & 0.55 & {0.70} & 0.50 & {0.75} & {0.95} & {0.95} & {0.95} & {0.95} & {0.25} & 0.05 & 0.05 & {0.65} \\ 
{Task 7}  & {0.35} & {0.35} & 0.15 & {0.85} & {0.40} & 0.05 & 0.55 & {0.65} & {1.00} & 0.90 & {1.00} & {1.00} & {0.00} & {0.00} & 0.10 & {0.55} \\ 
{Task 8}  & 0.40 & {1.00} & {0.80} & 0.50 & {0.70} & 0.15 & {0.80} & 0.45 & 0.80 & {0.95} & {1.00} & 0.85 & {0.05} & 0.00 & 0.05 & {0.20} \\ 
Task 9  & {0.90} & 0.60 & {1.00} & 0.90 & {0.50} & 0.00 & {0.80} & {0.80} & 0.45 & {0.85} & {1.00} & 0.95 & {0.10} & 0.00 & 0.00 & {0.10} \\ 
{Task 10} & 0.80 & {0.85} & 0.80 & {1.00} & {0.40} & 0.00 & {0.40} & 0.35 & 0.15 & {0.70} & 0.80 & {0.90} & 0.30 & {0.35} & 0.15 & {0.40} \\ \bottomrule
\end{tabular}}
\label{tab:language_condition}
\end{table*}


\myheading{Baselines \& Compared Methods.} We adopt \textit{OpenVLA}~\cite{DBLP:OpenVLA2024} as the initial baseline, as it offers a strong visuomotor policy but is computationally challenging. 
Our study emphasizes visual tokenization strategies, which are orthogonal to the choice of pretrained visuomotor backbones; accordingly, we limit our comparisons to OpenVLA as a representative reference. 
In summary, OpenVLA provides a fair starting point for evaluating our object- and object-relation–based approaches for the following comparison: \textit{OpenVLA}~\cite{DBLP:OpenVLA2024}: dense-token policy baseline with full token input, \textit{Object-Centric Slots (OC)}: independent object representations without explicit modeling of interactions (i.e. \model without Relation Encoder), and \textit{Object-Relation-Centric Slots (ORC)}: our proposed object-relational tokenization that captures object and object-context interactions (i.e. \model). 

\myheading{Evaluation Configuration.} 
\label{sec:eval-config}
We use 16 slots for L-Goal (simpler layouts) and 24 slots for L-Object, L-Spatial, and L-Long. After filtering, only 4 task-relevant object slots are retained, with relation slots matched to the initial object slots. This setup enables fair comparison without filtering and highlights the limitation of L-Long (up to 29 objects), which cannot be fully covered unless all task-relevant slots are retained. Training is performed with a batch size of 64 on 3 A100 GPUs for 50k iterations.

\myheading{Evaluation Protocol.} 
Each method is trained jointly across tasks within a subset and evaluated on 20 rollouts per task. We report {average success rate} per subset, along with ablation studies on slot scaling and temporal tracking.

\begin{table}[t!]
\centering
\caption{Ablation study on number of object tokens.}
\vspace{1em}
\begin{tabular}{l|ccc}
\toprule
\textbf{Method} & \textbf{4 tokens} & \textbf{8 tokens} & \textbf{16 tokens} \\
\midrule
OC                 &  {0.77}              &       {0.65}          &         {0.77}        \\
ORC                &     {0.86}           &        {0.74}         &       0.72          \\
\bottomrule
\end{tabular}
\label{tab:token_scaling}
\end{table}

\begin{figure*}[!t]
    \centering
\vspace{0.15cm}
    \includegraphics[width=0.9\linewidth]{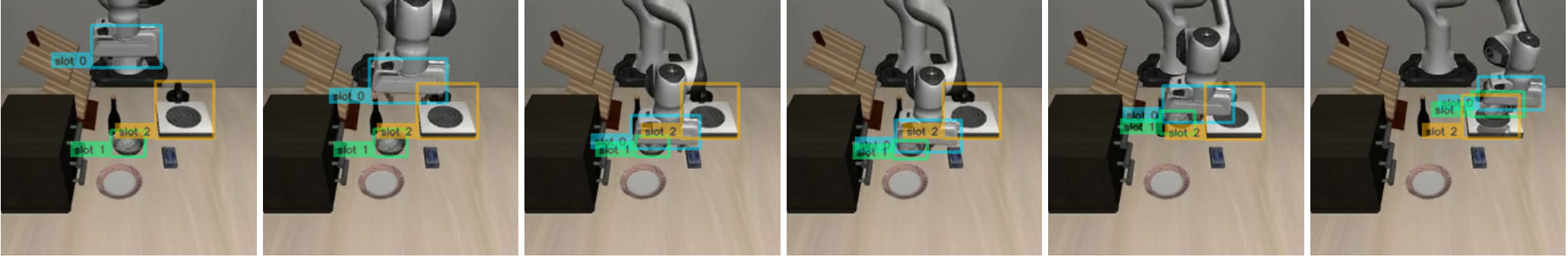}
    \vspace{0.12in}
    \caption{{Trajectory demonstration in simulation from exocentric views.} Task query: ``Put the bowl on the stove''.} 
    \label{fig:trajectory_demo}
    \vspace{-0.5em}
\end{figure*}

\subsection{Main Results}
\vspace{-0.5em}

Table~\ref{tab:main_results} reports success rates across the four subsets in LIBERO+. We can observe that OC yields comparable or slightly higher averages than OpenVLA in L-Goal (0.77) and L-Object (0.90), suggesting that compact object-centric slots can support reasoning when tasks are primarily object-driven with simple layouts. ORC provides additional gains in L-Goal (0.86) and L-Object (0.91) by explicitly encoding relational structure, though its advantage is less clear in L-Spatial (0.60) and it underperforms OpenVLA in L-Long (0.31 vs. 0.56). Thus, while relation-centric modeling is beneficial for object-focused tasks, its impact is more limited for spatially complex and long-horizon settings. 
A key dimension of this comparison is the trade-off between token number and computational cost. Slot-based models reduce the token count by more than an order of magnitude compared to OpenVLA (e.g., 4–28 slots vs. 256 dense tokens). This efficiency helps explain their competitiveness on L-Goal and L-Object, since these subsets involve only a few active objects where compact slots capture the essential cues without wasted computation. However, the same compactness becomes a limitation on L-Spatial and L-Long, where the available object slots (4) are so much fewer than the number of distinct objects (up to 29 in L-Long) that would also necessitate better relational modeling. Otherwise, not all objects can be represented, and fine-grained spatial layouts remain hard to capture, leading to degraded performance compared to the dense baseline. The key takeaway is that slot-based tokenization offers strong efficiency gains but its effectiveness depends on task complexity. OC and ORC perform well on simpler tasks (L-Goal, L-Object) by focusing on task-relevant objects and gripper positions. OC, however, struggles with changing layouts and many objects (L-Spatial, L-Long), failing especially when filtered to only four slots. ORC better captures relational reasoning but still falters when object sets grow large, as in L-Long. Thus, while compact slots are efficient and competitive, scaling to complex layouts, long horizons remains an open challenge.

\begin{table}[!t]
\centering
\caption{Ablation study on the effect of temporal consistency.}
\vspace{1em}
\begin{tabular}{l|cc}
\toprule
\multirow{2}{*}{\textbf{Method}} & \multicolumn{2}{c}{\textbf{Temporal Consistency}} \\ 
 & \textbf{\xmark} & \textbf{\cmark} \\
\midrule
OC   &                            0.38     & {0.77}        \\
ORC &                           0.40  & {0.86}             \\
\bottomrule
\end{tabular}
\label{tab:temporal_tracking}
\end{table}

\subsection{Task-Aware Slot Filtering Ablation}
\noindent We ablated the slot filtering in Sec.~\ref{subsec:task-centric} with two models:
\begin{itemize}
\item \textbf{With Slot Filtering (default):} only a small number of slots are passed downstream. OC retains 4 object slots; ORC uses 4 object slots plus 16--24 relation slots depending on the subset (Sec.~\ref{sec:eval-config}).
\item \textbf{Without Slot Filtering:} all extracted slots are used (L-Goal 16, L-Object 24, L-Spatial 24, L-Long 24).
\end{itemize}

Table~\ref{tab:language_condition} shows that both OC and ORC perform well on subsets of simple layouts (L-Goal, L-Object), where reasoning depends on picking and placing a few task-relevant objects. Performance drops on L-Spatial and L-Long, which involve changing layouts and many more objects. In L-Long, OC remains somewhat feasible with 24 slots (0.11) but collapses when filtered to 4 (0.07), showing the limits of non-relational modeling. ORC, though better for relational reasoning, also suffers with only 24 slots when scenes contain up to 29 objects, as its module cannot recover unless task-relevant slots are consistently included. We can observe here that by filtering for irrelevant objects while focusing on relevant ones, ORC can perform more decently on L-Long. Overall, this indicates that slot-based models are effective in object-centric settings with simple object layouts, but scaling to complex, long-horizon tasks would require both relevant object coverage and strong relational reasoning.
\vspace{-0.05in}
\subsection{Ablations and Insights}
Our ablation study is conducted on L-Goal for both OC and ORC with task-aware slot filtering enabled.

\textbf{Number of object slots.}  
Table~\ref{tab:token_scaling} reports results on L-Goal when varying the number of object slots. For OC, performance is the same at 4 and 16 tokens (0.77) but drops at 8 tokens (0.65), suggesting that simply scaling slot numbers does not yield consistent gains, but rather adds noise to the system because of increased number of irrelevant objects. For ORC, performance is strongest at 4 tokens (0.86) and decreases slightly as more slots are added (0.74 at 8, 0.72 at 16). This indicates that relational encoding enables the use of a small slot budget, but adding more slots may dilute relational grounding or introduce distractors in simpler tasks like L-Goal, where only a few objects are active. 

\textbf{Temporal consistency.}  
Table~\ref{tab:temporal_tracking} evaluates the effect of temporal consistency in action decoding. Without it, both OC (0.38) and ORC (0.40) degrade sharply, even falling below the dense OpenVLA baseline (0.77 on L-Goal). With consistency, performance improves markedly (0.77 for OC, 0.86 for ORC). The larger gap for OC shows its sensitivity to identity drift, while ORC is more robust but still affected. These results highlight that slot-based models depend not only on representation quality but also on stable slot identities; without consistency, the Action Decoder must re-ground objects at every frame, undermining task success.

\subsection{Qualitative Results.}
We provide qualitative analysis on the L-Goal task ``Put the bowl on the stove''.  

\textbf{Slot decomposition.}  
Fig.~\ref{fig:slot_decomposition} shows the decomposition of an input image into a set of slots. Each slot specializes to a distinct object or region, with high attention weights assigned to task-relevant objects such as the gripper (slot 0), the bowl (slot 1) and the stove (slot 2). Irrelevant background areas are assigned negligible weights, indicating that the slot-based encoder is able to factorize the scene into interpretable entities while filtering out distractors.  

\textbf{Trajectory prediction with stable slots over time.}  
Fig.~\ref{fig:trajectory_demo} shows the predicted trajectories of task-relevant objects over time. Slots maintain consistent identities across frames (e.g., slot~0/1 track the gripper and bowl, slot~2 the stove), demonstrating temporal consistency for both dynamic and static objects. This stability enables the action decoder to generate coherent predictions as the robot grasps the bowl and places it on the stove, highlighting how slot-based representations capture object semantics while preserving identity over time.

\section{Discussion}

As the first investigation into slot-based VLA, our results highlight both the potential and the limitations of slot-based tokenization for visuomotor reasoning. Our experiments show that compact slots in both OC and ORC can reduce token count and FLOPs by $3$–$4\times$ compared to dense baselines, while maintaining competitive accuracy on simpler tasks (L-Goal, L-Object), highlighting an efficiency–performance trade-off when only a few objects matter. However, scaling slot numbers does not guarantee gains: for OC, accuracy stagnates or even drops with more slots, while for ORC, performance weakens as additional slots dilute relational grounding, suggesting that compactness itself is often advantageous. Temporal consistency also proves critical; without it, both OC and ORC fall much lower, showing that stable slot identities are essential for effective action decoding.

\section{Conclusion}
We have introduced \textbf{LIBERO+}, a benchmark enriched with object-centric annotations, and \model, a slot-based framework for structured visuomotor control. By leveraging compact object and relation slots, our approach reduces token usage by an order of magnitude compared to dense-token baselines, while maintaining competitive task performance. The results highlight that object–relation–centric tokenization offers a promising balance of efficiency, interpretability, and effectiveness for multitask robotic manipulation. At the same time, challenges remain in scaling to long-horizon and cluttered settings, where slot coverage and temporal consistency become critical. We hope LIBERO+ and \model will provide a foundation for future work on structured, efficient, and explainable embodied AI.

\myheading{Limitations.} While \model leverages relations effectively in simple subsets of limited object layouts, it struggles to scale in more complex scenarios, suggesting the need for more fine-grained relational modeling, with more relation slots in the future when more compute available. Second, slot construction depends on full supervision with bounding boxes and masks, which may reduce applicability in real-world settings. Third, relation-centric slots are only implicitly learned through cross-slot attention and visual features, leaving room for more explicit relational grounding.

\myheading{Broader Impacts.} 
Slot-based representations under \model offer interpretability and efficiency, making them appealing for safer and more trustworthy embodied AI. Compact tokens may allow robots to operate under limited compute budgets while still reasoning over relevant entities. More broadly, improving transparency in object–relation grounding can foster safely structured human-AI collaboration.

\vspace{-0.05in}
\bibliographystyle{IEEEtran}
\bibliography{main}

\end{document}